\documentclass[letterpaper]{article} % DO NOT CHANGE THIS
\usepackage{aaai2026}  % DO NOT CHANGE THIS
\usepackage{times}  % DO NOT CHANGE THIS
\usepackage{helvet}  % DO NOT CHANGE THIS
\usepackage{courier}  % DO NOT CHANGE THIS
\usepackage[hyphens]{url}  % DO NOT CHANGE THIS
\usepackage{graphicx} % DO NOT CHANGE THIS
\usepackage{natbib}  % DO NOT CHANGE THIS AND DO NOT ADD ANY OPTIONS TO IT
\usepackage{caption} % DO NOT CHANGE THIS AND DO NOT ADD ANY OPTIONS TO IT
\usepackage{algorithm}

\usepackage{amsmath} % 提供 \operatorname 等数学命令
\usepackage{makecell}
\usepackage{bm}
\usepackage{algpseudocode}
\usepackage[table]{xcolor}
\usepackage{amsmath}
\usepackage{booktabs}
\usepackage{amssymb}
\usepackage{tcolorbox}
\usepackage{xcolor}
\usepackage{enumitem}
\usepackage{amsfonts}
\usepackage{multirow}

\tcbuselibrary{skins,breakable}

\usepackage{newfloat}
\usepackage{listings}
\DeclareCaptionStyle{ruled}{labelfont=normalfont,labelsep=colon,strut=off} % DO NOT CHANGE THIS
\floatstyle{ruled}
\newfloat{listing}{tb}{lst}{}
\floatname{listing}{Listing}
\title{Cost-Effective Communication: An Auction-based Method for  \\ Language Agent Interaction}
\author{
Yijia Fan\textsuperscript{\rm 1}\equalcontrib,
Jusheng Zhang\textsuperscript{\rm 1}\footnotemark[1],
Kaitong Cai\textsuperscript{\rm 1},
Jing Yang\textsuperscript{\rm 1},\\
Chengpei Tang\textsuperscript{\rm 1}\footnotemark[2],
Jian Wang\textsuperscript{\rm 2},
Keze Wang\textsuperscript{\rm 1,3}\thanks{Corresponding author: kezewang@gmail.com and tchengp@mail.sysu.edu.cn}
}

\affiliations{
\textsuperscript{\rm 1}Sun Yat-sen University\\
\textsuperscript{\rm 2}Snap Inc.\\
\textsuperscript{\rm 3}Guangdong Key Laboratory of Big Data Analysis and Processing\\
}

\usepackage{bibentry}
\begin{document}

\maketitle

\begin{abstract}
Multi-agent systems (MAS) built on large language models (LLMs) often suffer from inefficient ``free-for-all'' communication, leading to exponential token costs and low signal-to-noise ratios that hinder their practical deployment. We challenge the notion that more communication is always beneficial, hypothesizing instead that the core issue is the absence of resource rationality. We argue that ``free'' communication, by ignoring the principle of scarcity, inherently breeds inefficiency and unnecessary expenses. To address this, we introduce the \textbf{D}ynamic \textbf{A}uction-based \textbf{L}anguage \textbf{A}gent (\textbf{DALA}), a novel framework that treats communication bandwidth as a scarce and tradable resource. Specifically, our DALA regards inter-agent communication as a centralized auction, where agents learn to bid for the opportunity to speak based on the predicted \textit{value density} of their messages. Thus, our DALA intrinsically encourages agents to produce concise, informative messages while filtering out low-value communication.
Extensive and comprehensive experiments demonstrate that our economically-driven DALA achieves new state-of-the-art performance across seven challenging reasoning benchmarks, including 84.32\% on MMLU and a 91.21\% pass@1 rate on HumanEval. Note that this is accomplished with remarkable efficiency, i.e., our DALA uses only 6.25 million tokens, a fraction of the resources consumed by current state-of-the-art methods on GSM8K. Further analysis reveals that our DALA cultivates the emergent skill of strategic silence, effectively adapting its communication strategies from verbosity to silence in a dynamical manner via resource constraints. Our code and updates are available at \url{https://github.com/waltstephen/Cost-Effective-Communication}.
%thereby learning not just what to say, but also when to be quiet
\end{abstract}

\section{Introduction}

The research for Large Language Models (LLMs)\citep{LLM1,Z2} is rapidly shifting from single-agent\citep{singleagent,singleagent2} solvers to complex Multi-Agent Systems (MAS)\citep{mas1,Z6,Z7,fan-etal-2025-ccg}, which hold the promise of tackling problems beyond the reach of any individual model. This collaborative approach has shown initial success in domains like complex reasoning\citep{li2025lion} and software development\citep{mas2_code}, where agents can refine solutions by pooling distributed knowledge and engaging in iterative dialogue. 
However, these advancements often rely on ``free-for-all'' communication protocols, which permit agents to broadcast information at a low cost. This architectural reliance leads to an exponential increase in token consumption, inundating agents with low-value, verbose exchanges. Existing methods, such as PHP\citep{PHP} and DyLAN\citep{Dylan1}, despite their performance gains, overlook the resource cost of communication, resulting in high operational overhead and a low signal-to-noise ratio. For instance, on the GSM8K benchmark\citep{gsm8k}, these methods consume tens of millions of tokens, failing to achieve efficient deployment. In contrast to the current research that has predominantly focused on enhancing individual agent capabilities\citep{individualagent2,Z9,li2025cogvla,danet} or simple message pruning, the optimal design of communication protocols, i.e., shaping \textit{how} and \textit{when} agents should communicate, might also be beneficial and promising.

\begin{figure*}[t]
    \centering
    \includegraphics[width=\textwidth]{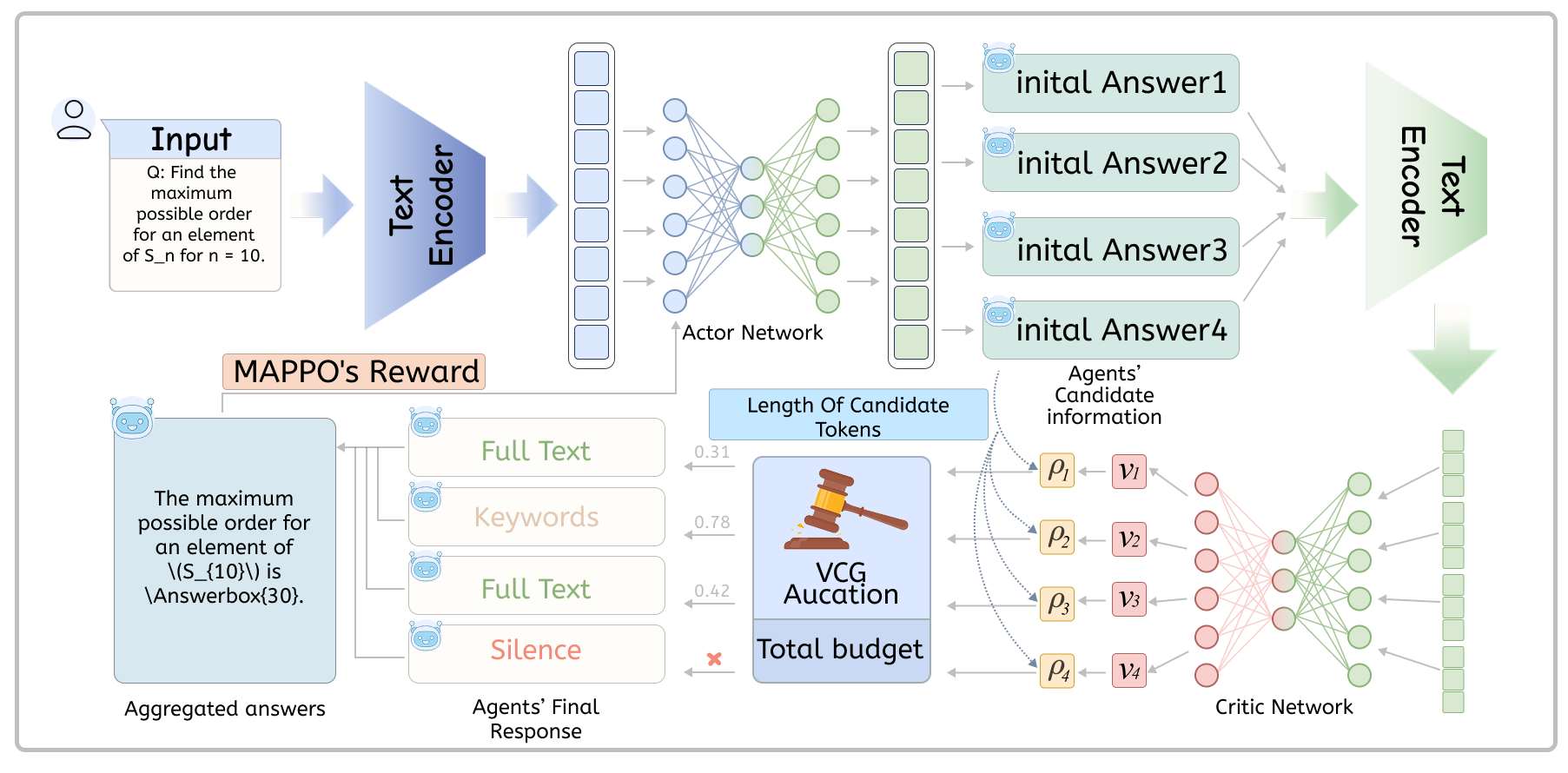}
    \caption{An overview of our DALA. Specifically, an Actor Network generates candidate messages, and a Critic Network computes their value density ($\rho_{i}$) based on utility ($V_{i}$) and cost. This density serves as a bid in a budget-constrained VCG auction that allocates speaking rights. The entire actor-critic system is trained with MAPPO to maximize a reward signal combining task success and auction costs, thereby learning an efficient communication strategy.}
    \label{fig:enter-label}

\end{figure*}

To achieve this objective, we challenge a fundamental question: the bottleneck in MAS performance is not a lack of communication, but the absence of \textit{resource rationality} in its budget management. We hypothesize that by treating communication bandwidth as a scarce, common resource and compelling agents to operate within a simulated market, the MAS can self-organize to prioritize concise, high-value information exchange. The core insight motivating this hypothesis is that ``free'' communication, by ignoring resource scarcity, breeds inefficiency; introducing incentive mechanisms can cultivate emergent rational intelligence, thereby balancing performance with efficiency of the MAS.

To validate this hypothesis, we introduce the \textbf{Dynamic Auction-based Language Agent (DALA)}, a novel framework that integrates principles of scarce resource allocation. Specifically, our DALA redefines inter-agent communication as a centralized auction process, such that agents must ``pay to speak'' and the entire framework is trained via the MAPPO strategy\citep{MAPPO}. Agents formulate bids based on the predicted value density of their potential messages, which mechanistically incentivizes them to generate concise and informative statements to win the auction. This mechanism intrinsically prunes low-value chatter, fostering a more deliberate and efficient conversational environment.

%, making it the first framework to truly ``teach'' a MAS when to be quiet.

In summary, our \textbf{main contributions} are: i) We introduce the principle of resource rationality to the field of MAS, identifying the management of available tokens in communication as a scarce resource, rather than its volume, as the key to unlocking both performance and efficiency; ii) We present DALA, a novel framework with a dynamic auction mechanism, compelling agents to evaluate the value density of their messages and bid for the opportunity to speak; iii) We empirically demonstrate that our DALA not only achieves new state-of-the-art performance with a fraction of the token cost but also cultivates the critical emergent skill of strategic silence, effectively teaching a MAS when to be quiet.
Extensive and comprehensive experiments demonstrate that our DALA yields significant improvements in both performance and efficiency, further establishing a new state-of-the-art across seven challenging reasoning benchmarks. %Specifically, it achieves 84.32\% accuracy on MMLU\citep{MMLU} and a 91.21\% pass@1 rate on HumanEval\citep{humaneval}. More importantly, its performance is accompanied by remarkable efficiency: on GSM8K, DALA consumes only 6.25 million tokens, a fraction of the 26 million tokens used by other high-performance MAS. %Further analysis reveals that agents in DALA learn to adapt their strategies based on resource constraints and information value, shifting from verbosity to silence.

\section{Related Work}
\label{sec:related_work}

\textbf{The Evolution of Multi-Agent Communication.}
The challenge of efficient communication in Multi-Agent Systems (MAS) is not new, but has been amplified by the token-based economics of modern Large Language Models (LLMs). Early MAS research, guided by Speech Act Theory \citep{searle1969speech,10504912,10.1145/3664647.3681212}, led to the development of standardized Agent Communication Languages (ACLs) to ensure semantic clarity \citep{fipa2002acl}. The subsequent Contract Net Protocol (CNP) \citep{smith1980contract} established a pragmatic negotiation framework for task allocation, but its designers already recognized that unconstrained communication could lead to prohibitive overhead. With the rise of deep reinforcement learning, the focus shifted from fixed protocols to learning emergent communication strategies. However, "free-for-all" communication was found to introduce noise often and hinder collaboration \citep{sukhbaatar2016learning}. This has prompted research on communication efficiency, focusing on three aspects: topology pruning \citep{jiang2018learning}; gated pruning \citep{DingLA,fan-etal-2025-towards}; and representation learning \citep{MACCA1,MACCA2}. While effective, they often lack a unified, principled framework to explicitly weigh the benefits of a message against its quantifiable cost.

\textbf{Economic Mechanisms and Information-Theoretic Principles.}
When communication bandwidth is a scarce resource, its allocation becomes an economic problem. Auction theory provides a rigorous framework for such problems, with the Vickrey-Clarke-Groves (VCG) mechanism being particularly notable \citep{vickrey1961counterspeculation,Z10}. VCG is incentive-compatible, encouraging truthful bidding, and maximizes social welfare, ensuring resources are allocated to agents who value them most. Its successful application in complex scenarios, from radio spectrum allocation \citep{cramton2006combinatorial} to Multi-armed Bandit System \citep{MABVCG}, validates its suitability for managing scarce communication resources in MAS. Our work operationalizes this by introducing the concept of \textit{value density} ($\rho = \text{value} / \text{cost}$), which determines an agent's bid. This concept deeply resonates with the Information Bottleneck (IB) principle \citep{tishby2000information}. DALA puts the information bottleneck principle into practice by training agents to compress their private knowledge into high-value density messages that contribute the most to the team's tasks while having the lowest token cost for themselves.

\textbf{Communication Efficiency in LLM-based MAS.}
Recent MAS frameworks built on LLMs have achieved success in complex reasoning and software development \citep{zheng2023progressive, du2023improving, zhang2023building,Z11}. However, their performance often comes at the cost of exponential token consumption due to ``free-for-all" communication protocols, resulting in a low signal-to-noise ratio. While some works \citep{mascom1,mascom2} have started to address this, for instance by pruning communication links based on random graph topologies like in AgentPrune-R \citep{zhao2023agentprune}, these methods lack an intelligent, value-driven basis for their pruning decisions. By integrating multi-agent reinforcement learning with the VCG auction mechanism, our DALA creates a communication market to force agents to bid for the opportunity to speak, thereby cultivating a resource rationality characterized by ``strategic silence" for a balance between performance and cost.
 % 如果文件名有空格，建议去掉或用下划线代替
\section{Methodology}

%In this section, we introduce the theoretical foundations of the \textbf{Dynamic Auction-based Language Agent (DALA)} framework. DALA reformulates multi-agent communication as a constrained dynamic resource allocation problem, employing market-based mechanisms to achieve globally optimal communication efficiency through individual utility maximization.

\textbf{Problem Formulation.}
%Our DALA models the MAS as a partially observable stochastic game, where communication bandwidth is a scarce shared resource that agents compete for through strategic bidding.
The learning objective of our DALA is to find the optimal joint policy parameters $\Theta^* = \{\theta_1^*, \dots, \theta_N^*\}$ that solve:
\begin{equation}
\begin{split}\label{eq:1}
    \Theta^* =& \arg\max_{\Theta}  \mathbb{E}_{\tau \sim \Pi_\Theta} [\mathcal{R}(\tau)] \\
    \text{s.t.} \quad & \mathbb{E}_{\tau \sim \Pi_\Theta} \left[\sum_{t=1}^{T} \sum_{a_j \in W_t} L(m_j^{(t)})\right] \le B
\end{split}
\end{equation}
Note that the budget in Eqn.~(\ref{eq:1}) is an \textit{expectation constraint} (a soft constraint), whereas the per-round budget in the auction mechanism is a hard constraint, practically enforcing the soft constraint's optimization objective.

\subsection{Value-Based Bidding Mechanism}

We design a principled bidding mechanism where agents learn to evaluate the marginal utility of their messages via a learned value network and bid accordingly.
% ========== REPLACE the existing paragraph under Section 2.2 with this more specific one ==========
The value-based bidding process is initiated through a prompt annotation mechanism guided by the \textbf{Actor Network}. Upon receiving an input question, it is first tokenized. The \textbf{Actor Network} then processes this sequence of tokens and outputs a scalar importance weight for each token, effectively appraising the salience of each word in the input. A new, annotated prompt is then constructed by interleaving the original words with their computed importance weights. For instance, a prompt like ``Explain the Pythagorean theorem" might be transformed into ``Explain [0.3] the [0.2] Pythagorean [0.9] theorem [0.8]". This annotated prompt, along with a meta-instruction explaining that the bracketed values signify token importance, is then passed to the agent's core language model. This directs the LLM's attention to the most critical parts of the query, guiding it to generate a more focused and relevant initial candidate message, which forms the basis for the subsequent value evaluation and bidding process.

\textbf{Guiding Rational Bidding.}
The core design objective of our DALA is to guide agents to progressively learn an approximately rational bidding strategy through our proposed reward mechanism and learning algorithm. Ideally, an agent's bid can accurately reflect the expected marginal contribution of its candidate message to the team's task objective, thereby achieving efficient allocation of communication resources.
%Message-Conditioned Value Function $V_{\theta_i}^V$:
To evaluate the utility of a specific message, we learn a \textbf{Message-Conditioned Value Function} $V_{\theta_i}^V$ for each agent $a_i$. This function aims to predict the expected cumulative return for the team, given the current observation $o_i^{(t)}$, if a candidate message $m$ is generated and broadcast. It is computed as follows:
\begin{equation}
v_i(m, o_i^{(t)}) = g_V(\phi_M(m) \oplus \phi_O(o_i^{(t)}); \theta_i^V),
\end{equation}
where $\phi_M$ and $\phi_O$ are message and observation encoders, respectively, $\oplus$ is a multi-modal fusion operator (such as an attention mechanism), and $g_V$ is an MLP head. $v_i$ is essentially an action-conditioned value, similar to a Q-value.
To balance the absolute value of a message with its cost (length) during bidding, we define the normalized value density $\rho_i(m, o_i^{(t)})$ as:
\begin{equation}
\rho_i(m, o_i^{(t)}) = \frac{v_i(m, o_i^{(t)}) - \bar{v}_t}{\sigma_{v_t} + \epsilon} \cdot \frac{1}{L(m)},
\end{equation}
where $\bar{v}_t$ and $\sigma_{v_t}$ are the mean and standard deviation of the values of all candidate messages in the current auction round, used to dynamically normalize the value to a Z-score, and $\epsilon$ is a small stability term.

The agent $a_i$'s \textbf{Bidding Function} $b_i^{(t)}$ is directly determined by the value density of its message:
\begin{equation} \label{eq:bid}
b_i^{(t)} = \max(0, \rho_i(m_i^{(t)}, o_i^{(t)})).
\end{equation}
This ensures that only messages with an expected positive marginal utility (after normalization) enter the auction. Based on the magnitude of the value density, we design a tiered content output strategy: when $\rho_i \geq \tau_{full}$, a full, detailed message is generated (Full); when $\tau_{summary} \leq \rho_i < \tau_{full}$, a summary is generated (Summary); when $\tau_{keywords} \leq \rho_i < \tau_{summary}$, only keywords are output (Keywords); and when $\rho_i < \tau_{keywords}$, the agent chooses to remain silent (Silence), where $\tau_{full} > \tau_{summary} > \tau_{keywords} > 0$ are preset value density thresholds.

These thresholds effectively divide the positive value density range, enabling a four-level message generation strategy. After an agent generates a candidate message, our DALA computes its value density $\rho_i$. Based on where this value falls, a secondary prompt is used to format the final output. A value in the top third of the positive range prompts a \texttt{Full} text; the middle third prompts a \texttt{Summary}; and the lowest positive third prompts \texttt{KeyWords}. If the value density is not positive (i.e., $\rho_i \le 0$ as per Eqn.~(\ref{eq:bid}), the agent chooses to remain in \texttt{Silence}. For example: \textit{``Q: Answer the question about the origin of the Pythagorean theorem", A: Keywords: Multiple origins (China, Babylonian civilization), Geometric relationship (right triangle side length formula), Pythagorean proof (from experience to deduction), Mathematical foundation (wide application and far-reaching influence)."}

\subsection{The DALA Combinatorial Auction Mechanism}
To enable multiple agents to communicate in a single round, we upgrade the auction to a \textbf{Combinatorial Auction} and apply the VCG payment rule\citep{groves1973incentives}. Instead of selecting a single winner, the goal is to determine a winning \textit{set} of agents $W_t$ whose combined messages provide the highest total value to the team, without exceeding the round's budget. This is known as the \textbf{Winner Determination Problem (WDP)}\citep{WDP}.

\textbf{Winner Determination Problem (WDP).} Given the set of valid bids $\mathcal{B}_{\text{valid}}$, the auctioneer solves for the winning set $W_t$ that maximizes the sum of bids, subject to the round's budget constraint $B_{max}^{(t)}$. This is a classic 0/1 Knapsack Problem:
\begin{equation} \label{eq:wdp}
\scalebox{0.8}{ % 这里的 0.9 是缩放比例，你可以根据需要调整
  $ \displaystyle % 保持公式元素的默认大小
  W_t = \arg\max_{S \subseteq \mathcal{B}_{\text{valid}}} \left\{ \sum_{a_j \in S} b_j^{(t)} \right\} \quad \text{s.t.} \quad \sum_{a_j \in S} L(m_j^{(t)}) \le B_{max}^{(t)}
  $
}
\end{equation}
While the WDP is NP-hard in general, it can be solved efficiently for moderately sized agent groups using dynamic programming.

\textbf{VCG Payment in Combinatorial Auction.} The VCG mechanism ensures that bidding truthfully remains the dominant strategy for each agent. For each winner $a_j \in W_t$, its payment $p_j^{(t)}$ is the social cost or ``harm" its presence imposes on all other agents. This is calculated as the optimal total value the framework \textit{could have achieved without} agent $j$, minus the value that all \textit{other winners actually received} in the presence of agent $j$.
\begin{equation} \label{eq:vcg_payment_combinatorial}
p_j^{(t)} = \left( \max_{S \subseteq \mathcal{B}_{\text{valid}} \setminus \{j\}} \sum_{a_k \in S} b_k^{(t)} \right) - \left( \sum_{a_k \in W_t \setminus \{j\}} b_k^{(t)} \right)
\end{equation}
If $a_i$ does not win (i.e., $a_i \notin W_t$), its payment $p_i^{(t)}$ is 0.

\textbf{Token Budget Management Strategy} remains conceptually the same, but the cost per round $C^{(t)}$ is now the sum of token lengths from all winners in the set $W_t$. We employ a hierarchical control mechanism: first, a total episode budget $B_{episode}$; second, a dynamic round-level budget $B_{round}^{(t)} = \frac{B_{episode} - \sum_{k=0}^{t-1} C^{(k)}}{T - t + 1}$; and finally, an instantaneous hard cap $B_{max}^{(t)} = \min(B_{round}^{(t)}, B_{hard})$.
The winner ``pays" not with currency, but with the communication budget consumed by its message length. When the budget is tight or depleted, our DALA automatically compresses messages, forces silence, and issues budget warnings to agents to guide them toward generating shorter content.

\begin{algorithm}[t]
\caption{DALA Combinatorial Auction Mechanism}
\label{alg:DALA_combinatorial_auction}
\begin{algorithmic}[1]
\Require Action set $\{(m_i^{(t)}, b_i^{(t)})\}_{i=1}^N$, current budget $B_{max}^{(t)}$.
\State \textbf{Budget Filtering:} Create a set of valid candidates $\mathcal{B}_{\text{valid}} = \{(m_i^{(t)}, b_i^{(t)}) \mid L(m_i^{(t)}) \leq B_{max}^{(t)}\}$.
\State \textbf{Winner Determination:} Solve the WDP for $\mathcal{B}_{\text{valid}}$ to find the winning set $W_t$. \Comment{Solves Eqn.~(\ref{eq:wdp})}
\State Initialize payments $p_i^{(t)} \leftarrow 0$ for all agents $i$.
\State Let total value of the winning set be $V(W_t) = \sum_{a_k \in W_t} b_k^{(t)}$.
\For{each winner $a_j \in W_t$}
    \State Let $\mathcal{B}_{\text{valid}}' = \mathcal{B}_{\text{valid}} \setminus \{a_j\}$.
    \State Solve WDP for $\mathcal{B}_{\text{valid}}'$ to find the hypothetical winning set $W_t^{-j}$.
    \State Let the optimal value without $j$ be $V(W_t^{-j}) = \sum_{a_k \in W_t^{-j}} b_k^{(t)}$.
    \State Let value of other winners in the actual outcome be $V(W_t \setminus \{j\}) = V(W_t) - b_j^{(t)}$.
    \State \textbf{VCG Payment} $p_j^{(t)} \leftarrow V(W_t^{-j}) - V(W_t \setminus \{j\})$. \Comment{Calculates Eqn.~(\ref{eq:vcg_payment_combinatorial})}
\EndFor
\State Actual budget cost $C^{(t)} \leftarrow \sum_{a_j \in W_t} L(m_j^{(t)})$.
\State \textbf{Output:} Winning agent set $W_t$, their messages $\{m_j^{(t)}\}_{j \in W_t}$, total cost $C^{(t)}$, and individual payments $\{p_j^{(t)}\}_{j \in W_t}$.
\end{algorithmic}
\end{algorithm}

\textbf{Multi-Agent Policy Optimization.}
The dynamics and reward structure of our DALA form the basis for multi-agent reinforcement learning. We optimize the agent policy parameters $\Theta = \{\theta_i\}_{i=1}^N$ through interaction with the environment.
Given the sparsity and team-based nature of the reward $R(\tau)$, we adopt the \textbf{Multi-Agent Proximal Policy Optimization (MAPPO)} algorithm. For agent $a_i$, we define the policy ratio as:
\begin{equation}
r_i(t)(\theta_i) = \frac{\pi_{\theta_i}(a_i(t) \mid o_i(t))}{\pi_{\theta_i^{old}}(a_i(t) \mid o_i(t))}
\end{equation}
where $\theta_i^{old}$ are the policy parameters from the last update.

The MAPPO clipped objective function is:
\begin{equation}
\begin{split}
    L_i^{\text{CLIP}} &(\theta_i) = \mathbb{E}_t \Big[ \min \big(  r_i(t)(\theta_i) A_i(t), \\
    & \text{clip} (r_i(t)(\theta_i), 1-\epsilon, 1+\epsilon) A_i(t) \big) \Big]
\end{split}
\end{equation}
where $A_i(t)$ is the advantage function for agent $i$ at time step $t$, and $\epsilon$ is the clipping parameter.
The value function loss is:
\begin{equation}
\begin{split}
    L_i^{\text{VF}} & (\theta_i) = \mathbb{E}_t  \Big[ \max \Big(  (V_{\theta_i}(o_i(t)) - R_i(t))^2, \\
    & (\text{clip}(V_{\theta_i}(o_i(t)), V_{\theta_i^{old}}(o_i(t)) \pm \epsilon_{vf}) - R_i(t))^2 \Big) \Big]
\end{split}
\end{equation}
where $R_i(t)$ is the actual return, and $\epsilon_{vf}$ is the clipping parameter for the value function. The final MAPPO objective function combines the policy loss, value loss, and entropy regularization:
\begin{equation}
L_i^{\text{MAPPO}}(\theta_i) = L_i^{\text{CLIP}}(\theta_i) - c_1 L_i^{\text{VF}}(\theta_i) + c_2 S[\pi_{\theta_i}](o_i(t))
\end{equation}
where $c_1, c_2$ are weighting coefficients, and $S[\pi_{\theta_i}]$ is the policy entropy.

\subsection{Learning Functions and Reward Design}
To guide rational bidding in the new combinatorial context, we adapt the immediate reward function.
\begin{equation} \label{eq:reward}
r_i(t) = \alpha \cdot \Delta_{task}(t) - \beta \cdot p_i^{(t)} \cdot \mathbb{I}[a_i \in W_t]
\end{equation}
where:
 $\Delta_{task}(t)$ remains the marginal contribution of the team's joint action to the task progress, e.g., $\Delta_{task}(t) = G_t - G_{t-1}$.
 $p_i^{(t)} \cdot \mathbb{I}[a_i \in W_t]$ is the VCG payment for agent $a_i$ if it is in the winning set $W_t$. This penalty, calculated via Eqn.~(\ref{eq:vcg_payment_combinatorial}), serves as agent $i$'s individual communication cost and incentivizes truthful bidding on its marginal value.
 $\alpha$ and $\beta$ are hyperparameters to balance task rewards and communication costs.

\textbf{Gradual Learning Mechanism:}
 \textbf{Message Value Learning}: This part remains unchanged. The message-conditioned value network $V_{\theta_i}^V$ still learns to predict the expected return for a specific message by minimizing the TD error:
    \begin{equation}
    L_{value} = \mathbb{E} \left[ (v_i(m_i(t), o_i(t)) - R_i(t))^2 \right]
    \end{equation}
    This allows $v_i$ to form the basis for the agent's bid $b_i(t)$.
 \textbf{Bidding Strategy Learning}: Through the updated reward function (\ref{eq:reward}), agents learn to generate messages and bids that are valuable not only in isolation, but also to contribute positively to a high-value \textit{combination}, i.e., winning a spot in the auction at a reasonable cost $p_i^{(t)}$.

Regarding the feedback mechanism, the reward signal guides agents toward forming beneficial coalitions. If an agent is part of a winning set $W_t$ whose collective action yields a high positive contribution (i.e., a high $\Delta_{task}(t)$), the entire team receives a positive task-related reward. This reinforces the behavior of generating messages that are valuable in combination with others. Conversely, if the winning set's collective action leads to a low or negative contribution, the team reward is negative, which penalizes and suppresses the formation of such ineffective coalitions in the future. Crucially, agents that are not part of the winning set ($a_i \notin W_t$) do not receive a payment penalty $p_i^{(t)}$ and only experience the team-wide task reward. This dynamic teaches agents to remain silent when their potential message is unlikely to be part of a high-value, winning coalition. Through multiple rounds of optimization on this reward signal using the MAPPO algorithm, the agents collectively internalize a more sophisticated rational logic: ``my information should be part of a high-value coalition to be worth bidding on," thus collaboratively achieving the team's objectives while adhering to budget constraints.
\section{Experiments}

%\subsection{Experimental Setup}

\paragraph{Baselines.}
To comprehensively evaluate the performance of our proposed DALA, we compare it against a suite of strong baselines. These are categorized into two groups: i) single-agent methods, including Vanilla prompting, Chain-of-Thought (CoT)\cite{CoT}, Complex CoT\cite{complexcot}, and Self-Consistency (SC)\cite{sc}; ii) multi-agent system (MAS) methods, including PHP\cite{PHP}, LLM-Debate\cite{LLM-debate}, DyLAN\cite{Dylan1}, and AgentPrune-R\cite{agentPrune-R}, which represents the current state-of-the-art communication pruning method based on a random graph topology.

\paragraph{Datasets.}
Our evaluation spans seven challenging benchmarks across three core reasoning domains. For general reasoning, we use MMLU\cite{MMLU}. For mathematical reasoning, we employ a diverse set of five benchmarks: GSM8K\cite{gsm8k}, MultiArith\cite{MultiArith}, SVAMP\cite{SVAMP}, AQuA\cite{AQuA}, and the particularly challenging MATH-500\cite{MATH} subset. For code generation, we use HumanEval \cite{humaneval} to assess function-level programming capabilities. For benchmarks with established training sets, we utilize them directly. For those without a standard training set, we adopt a data optimization paradigm(Similar to AgentPrune-R\cite{agentPrune-R}). Specifically, we designate 2\% of the data from each benchmark as an optimization set to train the DALA agent's bidding strategy, which is based on value density. After this policy is optimized via the MAPPO algorithm and subsequently fixed, we perform the final performance evaluation on the remaining 98\% of unseen data.

\paragraph{Experimental Settings.}
Following \cite{agentPrune-R} to ensure a fair comparison of the communication frameworks, all experiments are conducted using \texttt{gpt-4-1106-preview}\cite{openai2024gpt4technicalreport} as the base large language model for all agents. For all tasks, our DALA uses different total budgets to control the communication cost (see our supplementary materials). For each task instance, essential information, such as distinct mathematical formulas, contextual facts, or required code snippets, is distributed among the different agents. This setup ensures that no single agent possesses sufficient information to solve the problem alone, making effective and efficient communication a prerequisite for success and thereby creating an ideal testbed for our DALA. For specific experimental hyperparameter settings, please refer to Appendix "Implementation Details and Hyperparameters". For the specific comparison models, benchmarks, and datasets, please refer to Appendix ``Benchmarks and Baseline Methods". All experiments are performed on NVIDIA A100 80GB. 

\subsection{Performance and Cost-Effectiveness}

\begin{table*}[t]
  \centering
  %––––– 基本排版微调 –––––%
  \footnotesize
  %––––– 自动缩放至整页宽 –––––%
  %\resizebox{\textwidth}{!}{%
  \begin{tabular}{|l||c|c|c|c|c|c|c|}
    \hline
    Method & MMLU &MultiArith & GSM8K&SVAMP &
    AQuA & HumanEval & MATH-500 \\
    \hline\hline
    \multicolumn{8}{|l|}{\textit{Single-Agent Methods}} \\ \hline
    Vanilla        & 82.14 & 93.15 & 85.40 & 87.18 & 70.34 & 71.68 & 73.72 \\
    CoT            & 82.65 (+0.51) & 94.79 (+1.64) & 87.17 (+1.77) & 88.32 (+1.14) &
                     73.91 (+3.57) & 75.52 (+3.84) & 75.18 (+1.46) \\
    ComplexCoT     & 83.78 (+1.64) & 95.86 (+2.71) & 87.62 (+2.22) & 90.17 (+2.99) &
                     77.58 (+7.24) & 74.94 (+3.26) & 76.85 (+3.13) \\
    SC             & 82.66 (+0.52) & 96.88 (+3.73) & 87.93 (+2.53) & 88.69 (+1.51) &
                     75.08 (+4.74) & 77.30 (+5.62) & 77.02 (+3.30) \\
    \hline
    \multicolumn{8}{|l|}{\textit{Multi-Agent Methods}} \\ \hline
    PHP            & 83.45 (+1.31) & 96.41 (+3.26) & 92.45 (+7.05) & 90.62 (+3.44) &
                     76.25 (+5.91) & 82.96 (+11.28) & 79.24 (+5.52) \\
    LLM-Debate     & 83.69 (+1.55) & 96.27 (+3.12) & 90.23 (+4.83) & 90.56 (+3.38) &
                     77.52 (+7.18) & 83.79 (+12.11) & 80.15 (+6.43) \\
    DyLAN          & 80.16 (-1.98) & 94.27 (+1.12) & 88.16 (+2.76) & 87.40 (+0.22) &
                     74.16 (+3.82) & 89.70 (+18.02) & 81.66 (+7.94) \\
    AgentPrune-R   & 83.94 (+1.80) & 96.30 (+3.15) & 95.83 (+10.43) & 91.68 (+4.50) &
                     78.60 (+8.26) & 90.30 (+18.62) & 82.81 (+9.09) \\
    \hline
    \rowcolor{gray!22}
    DALA (Ours) &
    84.32 (+2.18) &97.87 (+4.72) &96.18 (+10.78) &
    92.33 (+5.15) & 80.92 (+10.58) & 91.21 (+19.53) &
    83.42 (+9.70) \\
    \hline
  \end{tabular}%
    \caption{Overall performance comparison across seven reasoning benchmarks. 
  We report accuracy (\%) for all tasks except HumanEval, for which we report pass@1 (\%). 
 Values in parentheses show absolute gain over the Vanilla baseline.}
  \label{tab:main_performance_updated}
%  } % end resizebox

\end{table*}
Table \ref{tab:main_performance_updated} demonstrates that our DALA consistently establishes the new state-of-the-art across all seven benchmarks. On the MMLU general reasoning task, DALA achieves an accuracy of 84.32\%, outperforming all evaluated baselines. This pattern of superiority is particularly evident in the domain of mathematical reasoning. For instance, on GSM8K, DALA reaches 96.18\% accuracy, a 1.35-point improvement over the previous best, AgentPrune-R. Similarly, on the challenging HumanEval and MATH-500 benchmarks, DALA demonstrates its robust capabilities by achieving top scores of 91.21\% and 83.42\%, respectively. These results underscore DALA's consistent ability to foster more effective collaboration that translates directly into superior problem-solving accuracy, irrespective of the task domain. Please refer to our supplementary materials for model computational overhead. For a specific QA case analysis, please see the Appendix "Case Study Analysis".

\begin{table}[h!]
\centering
% 使用 resizebox 可以确保表格宽度不超过页面文本宽度
\resizebox{\linewidth}{!}{%
\begin{tabular}{l cc cc}
\toprule
& \multicolumn{2}{c}{MMLU} & \multicolumn{2}{c}{GSM8K} \\
\cmidrule(lr){2-3} \cmidrule(lr){4-5}
Method & Accuracy (\%) & Token Cons. & Accuracy (\%) & Token Cons. \\
\midrule
\rowcolor{lightgray} Ours (DALA) & 84.3 & $3.80 \times 10^5$ & 96.2 & $6.20 \times 10^6$ \\
AgentPrune-R     & 84.0          & $9.60 \times 10^5$          & 95.8          & $7.50 \times 10^6$ \\
LLM-Debate       & 83.7          & $1.50 \times 10^6$          & 90.2          & $2.20 \times 10^7$ \\
PHP              & 83.4          & $2.60 \times 10^6$          & 92.5          & $2.60 \times 10^7$ \\
DyLAN            & 80.2          & $1.20 \times 10^6$          & 88.2          & $1.40 \times 10^7$ \\
Vanilla          & 82.1          & $1.50 \times 10^5$          & 85.4          & $3.50 \times 10^6$ \\
\bottomrule
\end{tabular}%
}
\caption{Consolidated performance comparison on MMLU and GSM8K datasets. Our DALA, highlighted in gray, demonstrates superior or competitive performance with significantly better token efficiency.}
\label{tab:combined_results}
\end{table}
Beyond superior performance, DALA is designed for economic efficiency, achieving top-tier accuracy at a fraction of the communication cost. As illustrated in Figure \ref{tab:combined_results}, DALA consistently occupies the high-accuracy, low-cost quadrant across benchmarks. On MMLU, it consumes only $1.81 \times 10^5$ tokens, an order of magnitude less than methods like PHP ($2.60 \times 10^6$). The advantage is even more pronounced on GSM8K, where DALA's token usage ($6.25 \times 10^6$) starkly contrasts with costly approaches like DyLAN ($1.40 \times 10^7$). This efficiency is a direct result of DALA's ``Pay-to-Speak" mechanism, a value-density auction that intrinsically prunes low-value, verbose messages, establishing DALA as a more practical and scalable framework.

\subsection{Analyses of Emergent Communication Strategies}

To validate the core hypothesis of our DALA that an economic-driven communication market can cultivate sophisticated and efficient discursive strategies, we conduct a multi-faceted analysis of the agents' emergent behaviors. This investigation aims to look beyond aggregate task performance and elucidate the intelligent, fine-grained decision-making processes fostered by our auction mechanism. To analyze strategic adaptation, we established two distinct total budget conditions. These settings are benchmarked against the token consumption of a standard baseline (Vanilla, i.e., \texttt{gpt-4-1106-preview}), which uses approximately $1.5 \times 10^5$ tokens for the MMLU task. We therefore set one budget at $1 \times 10^5$ tokens to create a resource-scarce scenario that necessitates efficiency, and a second, larger budget of $1 \times 10^6$ tokens to represent a resource-abundant environment. Please note that the tokens budget here refers to the budget relative to the entire task design. Under each condition, we recorded the frequency of four key communication strategies.
To further evaluate the adaptability of the strategy, we operationally define information types within the MMLU task: \textbf{critical information} refers to reasoning steps or facts that are essential for deriving the correct answer, while \textbf{non-critical information} includes redundant statements, incorrect reasoning paths, or conversational filler; the specific classification is performed by an external adjudicator model, i.e., \texttt{openai/o3-2025-04-16}. 

\begin{figure}[!t]
    \centering
    \includegraphics[width=\linewidth]{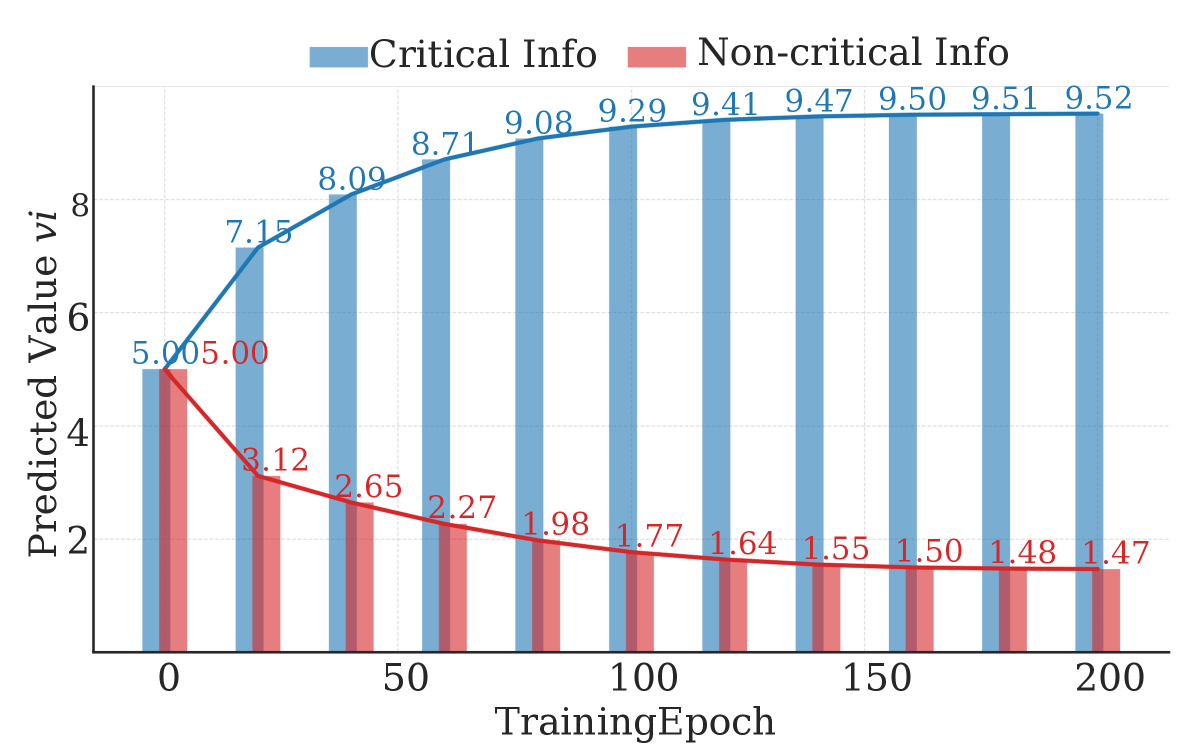}
    \caption{The learning curve of the average predicted value ($v_i$) on MMLU, showing the agent's internal value network rapidly learns to assign high value to critical information while devaluing non-critical information, with the valuation gap stabilizing as training converges around epoch 160.}
    \label{fig:value_curve}
\end{figure}

\begin{table}[t]
    \centering
    
    % 使用 resizebox 调整宽度，单栏或双栏皆宜
    \resizebox{\linewidth}{!}{
        \begin{tabular}{lcccc}
            \toprule
            \textbf{Domain} & \textbf{Full (\%)} & \textbf{Sum. (\%)} & \textbf{Key. (\%)} & \textbf{Silence (\%)} \\
            \midrule
            % --- 第一部分：Low Budget ---
            \multicolumn{5}{c}{\textbf{Low Budget ($1 \times 10^5$ Tokens)}} \\ 
            \midrule
            Humanities      & 5.1  & 24.9 & 30.1 & 39.9 \\
            Social Science  & 4.7  & 20.1 & 35.7 & 39.5 \\
            STEM            & 2.1  & 15.6 & 48.9 & 33.4 \\
            Other (General) & 6.9  & 28.3 & 29.5 & 35.2 \\
            
            \midrule
            % --- 第二部分：High Budget ---
            \multicolumn{5}{c}{\textbf{High Budget ($1 \times 10^6$ Tokens)}} \\ 
            \midrule
            Humanities      & 48.7 & 38.6 & 9.4  & 3.2 \\
            Social Science  & 45.9 & 42.1 & 8.9  & 3.0 \\
            STEM            & 30.6 & 42.1 & 21.1 & 6.3 \\
            Other (General) & 45.0 & 41.9 & 10.1 & 3.0 \\
            \bottomrule
        \end{tabular}
    }
    \caption{Comparison of agent communication strategies under different budget constraints. The table is split vertically into Low Budget (top) and High Budget (bottom) scenarios.}
    \label{fig:bandwidth_comparison}
\end{table}

The empirical results, presented visually, offer compelling validation of our DALA's underlying mechanics. The foundation of all strategic action is the agent's ability to appraise information. Figure \ref{fig:value_curve} presents the learning trajectory of this appraisal, showing that the value network rapidly learns to distinguish information types. Starting from near-identical valuations at epoch 0, the curves for critical and non-critical information diverge significantly, with the value delta stabilizing as the model converges around epoch 160. This learned ability to discern value is the crucial prerequisite that drives behavior in the communication market.

This internal valuation directly informs how agents adapt to resource constraints, as visualized in Figure \ref{fig:bandwidth_comparison}. Under the stringent $1 \times 10^5$ token budget (Figure \ref{fig:bandwidth_comparison}, bottom), agents across all domains are forced to be economical. They significantly curb the use of ``Full-text" messages (e.g., dropping to just 2.1\% in STEM) and increase their reliance on highly compressed ``keywords" (peaking at 48.9\% in STEM) and ``strategic silence" (39.9\% in Humanities). Conversely, with the generous $1 \times 10^6$ token budget (Figure \ref{fig:bandwidth_comparison}, top), agents become more expressive. ``Full-text" communication becomes a dominant strategy, especially in Humanities (48.7\%), while ``silence" diminishes to a negligible level across all domains (e.g., 3.0\% in Social Science). This stark contrast between the two conditions provides clear visual evidence that agents are not following a static policy, but are dynamically modulating their communication strategies based on the learned value of their information and the current state of the market. By analyzing the silence ratio of agents under different costs, the agents in our DALA have truly learned to ``speak less or not" when the total budget is small, and have truly learned \textbf{``when to shut up''}.

\textbf{Module Ablation.}
To demonstrate the detailed contributions of DALA's key components, we conduct a series of ablation studies on the MMLU and GSM8K benchmarks. We compare the full DALA model against several ablated variants, where each variant disables a single, critical mechanism. All other settings remain identical to our main experiments. The variants are as follows:
\textbf{w/o Value Learning}: Replaces the learned value function $V_\phi$ with a heuristic bid (e.g., proportional to message length), to verify that intelligent, learned valuation is essential for the auction's success. \textbf{w/o Value Density}: Bids are based on raw predicted value $v_i$, removing the $1/L(m)$ term. This tests the core hypothesis that optimizing for value-per-token, not absolute value, drives efficiency. \textbf{w/o Tiered Content}: Collapses the communication strategy to a binary choice (Full message vs. Silence), to assess the importance of adaptive granularity in managing the information-cost trade-off. \textbf{w/o Dynamic Budget}: Uses a fixed per-round budget instead of an adaptive one, to evaluate the benefits of a forward-looking resource allocation policy. \textbf{w/o Cost Penalty ($\beta=0$)}: Removes the communication cost from the reward function (Eq. 9), to demonstrate that an explicit economic penalty is necessary to enforce communicative discipline and prevent verbosity.
\begin{table}[t]
\centering
\scriptsize % 设置表格字体为 7pt (\scriptsize)
\begin{tabular}{lccc}
\toprule
\textbf{Variant} & \textbf{\makecell{MMLU \\ (Acc. \%)}} & \textbf{\makecell{GSM8K \\ (Acc. \%)}} & \textbf{\makecell{Avg. Tokens \\ ($\times 10^6$)}} \\
\midrule
\textbf{DALA (Full Model)} & \textbf{84.32} & \textbf{96.18} & \textbf{3.22} \\
\midrule
w/o Value Learning & 75.77 ($\downarrow$8.55) & 86.35 ($\downarrow$9.83) & 4.13 \\
w/o Value Density & 79.16 ($\downarrow$5.16) & 89.76 ($\downarrow$6.42) & 4.91 \\
w/o Tiered Content & 79.83 ($\downarrow$4.49) & 90.95 ($\downarrow$5.23) & 4.49 \\
w/o Dynamic Budget & 80.71 ($\downarrow$3.61) & 91.30 ($\downarrow$4.88) & 3.65 \\
w/o Cost Penalty ($\beta=0$) & 77.48 ($\downarrow$6.84) & 88.27 ($\downarrow$7.91) & 6.23 \\
\bottomrule
\end{tabular}
\caption{Ablation study on MMLU and GSM8K. We report accuracy (\%) and the average token consumption (Avg. Tokens) across both tasks. The results demonstrate that each component of our DALA contributes significantly to its overall performance and efficiency. The best performance and lowest token consumption are highlighted in \textbf{bold}.}
\label{tab:ablation_study}
\end{table}
The results of our ablation study (Table \ref{tab:ablation_study}) confirm the architectural integrity of our DALA, as removing any component causes a substantial decline in performance. The most pronounced degradation occurs with the ablation of \textbf{Value Learning}, causing an accuracy drop of up to 9.83 points. This validates our core hypothesis that an effective multi-agent system requires not just communication, but an intelligent appraisal of information's worth. Similarly, removing economic principles like \textbf{Value Density} and the \textbf{Cost Penalty} ($\beta=0$) leads to verbose, inefficient communication and information overload. This demonstrates the contribution of our DALA's components, i.e., value-based intelligence, economic discipline, and adaptive mechanisms like \textbf{Tiered Content} and a \textbf{Dynamic Budget}.

\section{Conclusion}

We presented DALA, a simple yet effective framework that recasts communication as a market-driven auction. By incentivizing agents to bid according to the \textit{value density} of their messages, DALA prunes low-value exchanges and encourages more deliberate communication. Experiments show that DALA achieves state-of-the-art performance on seven reasoning benchmarks with only a fraction of the token cost.

\section*{Acknowledgments}
This work was supported in part by the National Natural Science Foundation of China (NSFC) under Grant 62276283, in part by the China Meteorological Administration's Science and Technology Project under Grant CMAJBGS202517, in part by Guangdong Basic and Applied Basic Research Foundation under Grant 2023A1515012985, in part by Guangdong-Hong Kong-Macao Greater Bay Area Meteorological Technology Collaborative Research Project under Grant GHMA2024Z04, in part by Fundamental Research Funds for the Central Universities, Sun Yat-sen University under Grant 23hytd006, and in part by Guangdong Provincial High-Level Young Talent Program under Grant RL2024-151-2-11.

%    因为你的样式文件叫 aaai2026.bst，所以这里就用 aaai2026。
\bibstyle{aaai2026}

% 2. 指定 .bib 文件的文件名（注意：没有 .bib 后缀！）。
\bibliography{AnonymousSubmission/LaTeX/aaai2026}

\end{document}